\documentclass{article}

\usepackage{amsmath}
\usepackage{mathtools}
\usepackage{graphicx}

\usepackage[final]{neurips_2022}

\usepackage[utf8]{inputenc} 
\usepackage[T1]{fontenc}    
\usepackage{hyperref}       
\usepackage{url}            
\usepackage{booktabs}       
\usepackage{amsfonts}       
\usepackage{nicefrac}       
\usepackage{microtype}      
\usepackage{xcolor}         

\usepackage{xspace}
\usepackage{caption}
\usepackage{tikz}

\makeatletter
\DeclareRobustCommand\onedot{\futurelet\@let@token\@onedot}
\def\@onedot{\ifx\@let@token.\else.\null\fi\xspace}

\makeatother

\usepackage{graphicx}

\title{Region of Interest Detection in Melanocytic Skin Tumor Whole Slide Images}

\author{
  Yi Cui\thanks{Corresponding author.} \\
  Department of Economics\\
  University of North Carolina at Chapel Hill\\
  Chapel Hill, NC 27514 \\
  \texttt{yicui@unc.edu} \\
  \And
  Yao Li\\
  Department of Statistics \& Operations Research\\
  University of North Carolina at Chapel Hill\\
  Chapel Hill, NC 27514 \\
  \texttt{yaoli@email.unc.edu} \\
    \And
  Jayson R. Miedema\\
  School of Medicine\\
  University of North Carolina at Chapel Hill\\
  Chapel Hill, NC 27514 \\
  \texttt{jayson\_miedema@med.unc.edu} \\
    \And
  Sherif Farag\\
  School of Medicine\\
  University of North Carolina at Chapel Hill\\
  Chapel Hill, NC 27514 \\
  \texttt{sherif\_farag@med.unc.edu} \\
\And
  J.S. Marron\\
  Department of Statistics \& Operations Research\\
  University of North Carolina at Chapel Hill\\
  Chapel Hill, NC 27514 \\
  \texttt{marron@unc.edu} \\
  \And
    Nancy E. Thomas\\
  School of Medicine\\
  University of North Carolina at Chapel Hill\\
  Chapel Hill, NC 27514 \\
  \texttt{nancy\_thomas@med.unc.edu} \\
}

\begin{document}

\maketitle

\vspace{-10pt}
\begin{abstract}
Automated region of interest detection in histopathological image analysis is a challenging and important topic with tremendous potential impact on clinical practice. The deep-learning methods used in computational pathology help us to reduce costs and increase the speed and accuracy of regions of interest detection and cancer diagnosis. In this work, we propose a patch-based region of interest detection method for melanocytic skin tumor whole-slide images. We work with a dataset that contains 165 primary melanomas and nevi Hematoxylin and Eosin whole-slide images and build a deep-learning method. The proposed method performs well on a hold-out test data set including five TCGA-SKCM slides (accuracy of 93.94\% in slide classification task and intersection over union rate of 41.27\% in the region of interest detection task), showing the outstanding performance of our model on melanocytic skin tumor. Even though we test the experiments on the skin tumor dataset, our work could also be extended to other medical image detection problems, such as various tumors' classification and prediction, to help and benefit the clinical evaluation and diagnosis of different tumors.



\end{abstract}
\section{Introduction}

From the study of \citet{Pollack2011Melanoma2005}, melanocytic skin tumor that has metastasized (distant stage) or is thicker than 4.00 mm has a poor prognosis (5-year survival: 15.7\% and 56.6\%). Also, \citet{Gerger2005DiagnosticTumors, Argenziano2010Slow-growingStudy, Seidenari2012VariegatedMelanoma} show that established methods have been proved insufficient to diagnose melanoma at an early stage. One solution may be histopathological images, which have long been utilized in treatment decisions and prognostics for cancer. Deep learning-based predictors trained on annotated data and non-annotated data could be a potential efficient technology to improve the early detection and diagnosis of such tumors and overall survival rates of treatment.

\begin{figure}
\vspace{-10pt}
	\begin{center}             
		\includegraphics[width=1\linewidth]{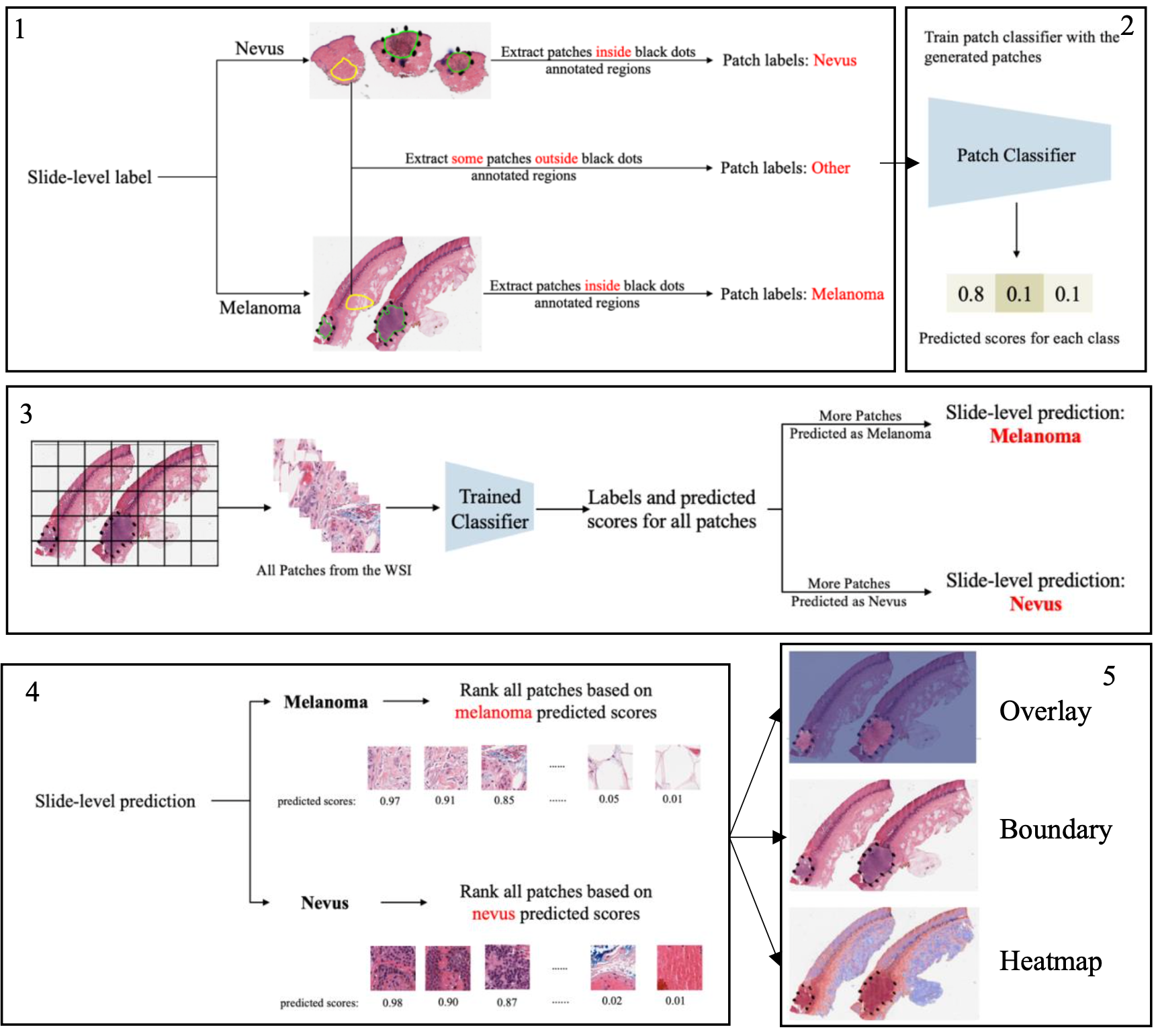}
	\end{center}
	\vspace{-10pt}
	\caption{\footnotesize Overview of the proposed detection framework. 1) Extract melanoma, nevus and other patches from training data. 2) Train a three-class patch classifier based on extracted patches. 3) For each slide, generate predicted scores for all patches and calculate patch as well as slide classification accuracy. 4) Rank all patches from a slide based on the corresponding predicted scores in the context of melanoma or nevus, depending on the slide classification result. 5) Generate visualization results based on predicted scores. }
	\label{fig:1}          
	\vspace{-0.6cm}
\end{figure}

The goal of this work is to develop a deep neural network-based Region of Interest (ROI) detection method that can precisely detect the ROIs in melanocytic skin tumors through Whole Slide Images (WSIs) and at the same time, diagnose the type of skin tumor ({\it melanoma} vs. {\it nevus}) accurately. However, most deep learning architectures cannot work on large WSIs directly due to their large sizes.
Recent studies \citep{Lerousseau2020WeaklySegmentation, Lu2021Data-efficientImages} have shown some progress in dealing with this issue by multiple instance learning. They propose to break down the large images into small patches and extract features from these patches, then slide-level features are generated by pooling the patch features together. Following that direction, we also break the large WSIs into patches and train our model with them. Besides, we leverage some partial information from annotations that are not made for ROI detection to help improve the performance of the model. Also, we show that the proposed method can work with various training set sizes by decreasing the size of the training set to different levels and testing the model performance on the same test set. Figure~\ref{fig:1} illustrates the overview of our method.

\section{Data and Methods}

	
\textbf{Data pre-processing.} WSIs (165 in total) from private melanocytic tumor dataset cohort and TCGA-SKCM cohort are randomly split and assigned to 80\% for training and 20\% for testing. Given a WSI, patches are extracted based on the slide-level label and annotations (see subfigure 1 of figure~\ref{fig:1}). If the slide-level label is nevus, all patches inside the annotated regions will be labeled as nevus. If the slide-level label is melanoma, all patches inside the annotated regions will be labeled as melanoma. Besides patches from annotated regions, some patches outside those regions are also extracted and labeled as other. The reason why we add the class of other is that if we just train a two class classifier model, it will wrongly classify some other patches as melanoma or nevus and those wrongly classified patches will also affect the slide classification result. However, since not all ROIs are annotated by pathologists, there could be melanoma and nevus patches outside annotated regions. To avoid labeling those patches as other, we manually extract patches of other classes from regions. To show that our method can work with small amount of training samples, we conduct the experiments with 80\% (107 WSIs), 60\% (80 WSIs), 40\% (54 WSIs), and 20\% (33 WSIs) of training samples, and test the models on the same test set (31 WSIs).


\textbf{Training data preparation and model.} 
A three-class patch classification model (PCLA-3C) is trained on the labeled patches with VGG16, see \citet{Simonyan2015VeryRecognition}, as base architecture. In the testing stage, all patches from a WSI will first be fed into the trained patch classifier. Ignoring patches predicted as {\it other}, slide-level prediction is done by majority vote based on patches predicted as {\it melanoma} and {\it nevus}. For a WSI classified as melanoma, all the patches from this slide will be ranked by melanoma predicted scores. Otherwise, all the patches will be ranked by nevus predicted scores. Three visualization maps are generated based on the predicted scores (See subfigure 5 of figure~\ref{fig:1}).

\textbf{Model assessment.} To evaluate the performance of ROI detection, the annotated ratio is measured to calculate Intersection over Union (IoU) for each slide. Given a slide, annotated ratio $\beta$ is calculated by the number of patches in the annotated regions divided by the number of patches extracted from the slide:
\begin{align*}\beta = \frac{A_p}{C_p},
\end{align*}
where $A_p$ is the number of patches in A (annotated region) and $C_p$ is the number of patches in C (WSI). Then, the top $n\beta$ patches based on predicted scores are classified as ROI, where $n$ is the total number of patches from a slide. The performance will be measured by the IoU, 
which compares the annotated regions and predicted ROI regions. Since the framework is patch-based, IoU is calculated by the number of patches in the intersection region (the region in both annotated and predicted regions) divided by the number of patches in the union of the annotated and predicted ROI regions:
\begin{align*}
    \text{IoU} = \frac{\underline{AB}_p}{\overline{AB}_p},
\end{align*}
where $\underline{AB}_p$ shows the number of patches in the region of $(A\cap B)$ and $\overline{AB}_p$ shows the number of patches in the region of $(A\cup B)$. A is annotated region and B is the predicted/highlighted region.

\vspace{-10pt}
\section{Results and discussion}
\vspace{-10pt}

\begin{table}
  \caption{Robustness performance of patch classification accuracy, slide classification and IoU by PCLA-3C and CLAM using different splits of the original training set. Since CLAM does not do patch classification, it does not have patch classification accuracy.}
  \label{tab:table1}
  \centering
  \resizebox{14cm}{!}{
  \begin{tabular}{c||cc|cc||cc|cc}
    \toprule
     & \multicolumn{4}{c||}{20\% split}  & \multicolumn{4}{c}{40\% split}               \\ \cmidrule(r){2-9}
     & \multicolumn{2}{c|}{PCLA-3C}  & \multicolumn{2}{c||}{CLAM} & \multicolumn{2}{c|}{PCLA-3C}  & \multicolumn{2}{c}{CLAM}               \\
    \midrule
             & Mean     & 95\% CI & Mean     & 95\% CI  & Mean     & 95\% CI & Mean     & 95\% CI \\
    \cmidrule(r){2-9}
    Patch classification accuracy& 0.6397&	[0.5193, 0.7601]& - & - & 0.7887&	[0.7536, 0.8238] & - & -\\
    Slide classification accuracy     & 0.7406&	[0.6627, 0.8185] & 0.6710&	[0.6386, 0.7033]&0.8430&	[0.8043, 0.8817] & 0.6976&	[0.6619, 0.7333] \\
    Intersection over Union     & 0.3026&	[0.2394, 0.3327]&	0.0427 &	[0.0342, 0.0512] &  0.3402&	[0.3057, 0.3784] & 0.0524&	[0.0297, 0.0751] \\
    \midrule
    & \multicolumn{4}{c||}{60\% split}  & \multicolumn{4}{c}{80\% split} \\ \cmidrule(r){2-9}
     & \multicolumn{2}{c|}{PCLA-3C}  & \multicolumn{2}{c||}{CLAM} & \multicolumn{2}{c|}{PCLA-3C}  & \multicolumn{2}{c}{CLAM}               \\
    \midrule
    & Mean     & 95\% CI & Mean     & 95\% CI & Mean     & 95\% CI & Mean     & 95\% CI\\
    \cmidrule(r){2-9}
    Patch classification accuracy& 0.8191 & [0.7766, 0.8616]& - & - & 0.8210& [0.7949, 0.8471] & - & - \\
    Slide classification accuracy     & 0.8721&	[0.8458, 0.8985]& 0.7097&	[0.6830, 0.7364]& 	0.8885&	[0.8607, 0.9163] & 0.7258&	[0.7117, 0.7399] \\
    Intersection over Union     & 0.3652&	[0.3369, 0.3934]& 0.0621&	[0.0428, 0.0814] & 0.3710	&[0.3335, 0.4084] & 0.1103	&[0.0529, 0.1677]\\
    \bottomrule
  \end{tabular}}
  \vspace{-0.5cm}
\end{table}

Two methods are tested on the melanocytic skin tumor dataset to finish ROI detection and slide classification: 1) CLAM (clustering-constrained attention multiple instance learning) from \citet{Lu2021Data-efficientImages}, 2) PCLA-3C (the proposed patch-based classification model)\footnote{\href{https://github.com/cyMichael/ROI\_Detection}{https://github.com/cyMichael/ROI\_Detection}}. Both methods are trained on the training set, and the performances on both training and testing sets are evaluated. The relatively low IoUs come from the fact that ground-truth labels are not available for evaluation. We can see more visualization results in Appendix.

We use CLAM and PCLA-3C by randomly selecting different proportions of original training data. We find that by utilizing 60\% of original training data, our accuracy is 0.8191 (95\% CI, 0.7766-0.8616) at the patch level, and accuracy is 0.8721 (95\% CI, 0.8458-0.8985) at the slide level, see table~\ref{tab:table1} for detailed values of each split. To calculate the 95\% CI and mean value for 60\% training data, for instance, we randomly split the training data by 10 times and then train the model on these independent subsamples. Our true testing data are kept unchanged since these data include true annotations. Table~\ref{tab:table1}’s results are based on the testing set, and we also test whether ROIs can be accurately detected in the testing set. As in the private melanocytic tumor dataset cohort, our pathology team annotated the TCGA-SKCM cohorts to mark ROIs. One figure in the appendix shows one representative sample of annotations (black dot regions) as well as without annotations. There is a high agreement between the predictions of the ROI and the PCLA-3C, demonstrating the generality of our automatic ROI detection. We perform CI estimation to enhance model performance. The unannotated and annotated models achieve an accuracy of 0.7406 (95\% CI: 0.6627-0.8185) at the slide level and an accuracy of 0.6397 (95\% CI: 0.5193-0.7601) at the patch level by only utilizing 20\% of original training samples. For the IoU value, we can see that we have a mean of 37.10\% for 80\% of training data, which is much higher than the results of CLAM. As in the PCLA-3C, the improvements in patch classification accuracy, slide classification accuracy and IoU show the importance of annotations in the training of deep learning classifiers for prediction. We also test whether melanoma, nevus or other status classifiers can directly predict certain tumor types. This is important as accurate tumor type is the clinical biomarker for future treatment. In summary, our deep-learning-based framework outperforms the state-of-the-art ROI detection methods like \citet{Lu2021Data-efficientImages}, which is quite crucial in medical imaging fields and related treatment recommendations.


\newpage
\bibliography{refer}

\clearpage

\appendix

\section{Appendix}
\subsection{Potential Negative Societal Impact}
To the best of our knowledge, we do not foresee many potential negative societal impacts to our work. Furthermore, the deep learning-based methods does not conduct any direct diagnosis but rather aids doctors with those ROI regions.

\subsection{Supplemental materials}

\textit{Computational configuration}. All analyses were used by Python. Images were analyzed and processed using OpenSlide. All the computational tasks were finished on Cluster with Linux (Tested on Ubuntu 18.04), NVIDIA GPU (Tested on Nvidia GeForce RTX 3090 on local workstations). NVIDIA GPUs supports were followed to set up and configure CUDA (Tested on CUDA 11.3) and the torch version should be greater than or equal to 1.7.1.

\begin{figure}
	\begin{center}             
		\includegraphics[width=1\linewidth]{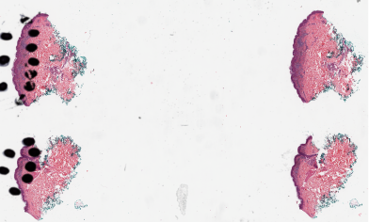}
	\end{center}
	\caption{One representative sample of annotations (black dot regions) as well as without annotations. Some ROIs on the right side are not annotated.}
	\label{fig:10}          
\end{figure}

\begin{figure}
\begin{minipage}[b]{1.0\linewidth}
  \centering
  \centerline{\includegraphics[width=5cm]{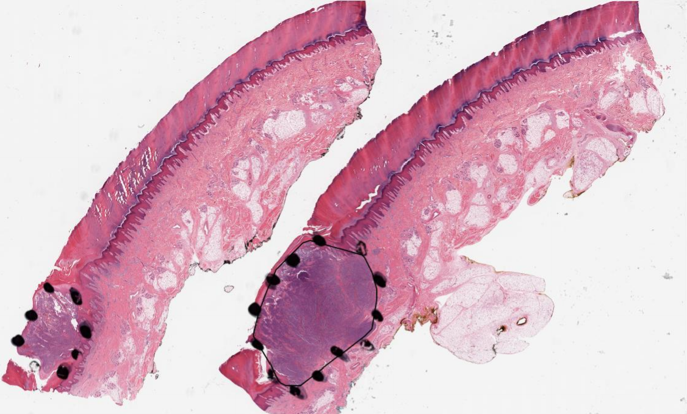}}
  \centerline{(a) Boundary of the largest predicted ROI region}\medskip
\end{minipage}
\begin{minipage}[b]{.48\linewidth}
  \centering
  \centerline{\includegraphics[width=5cm]{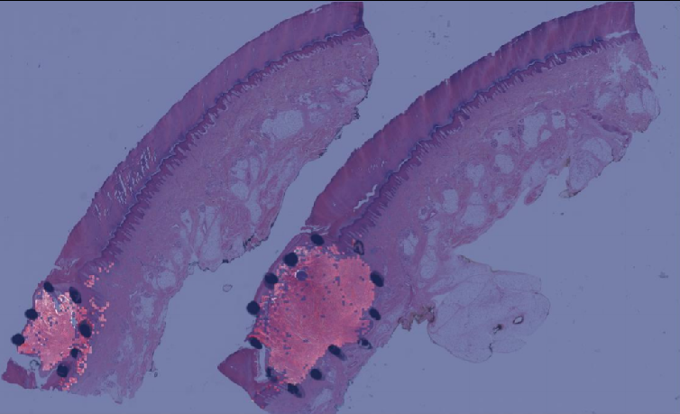}}
  \centerline{(b) Overlay}\medskip
\end{minipage}
\hfill
\begin{minipage}[b]{0.48\linewidth}
  \centering
  \centerline{\includegraphics[width=5cm]{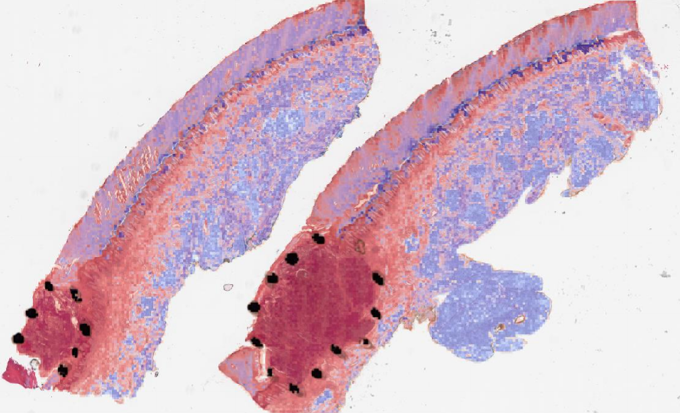}}
  \centerline{(c) Heatmap}\medskip
\end{minipage}
\vspace{-10pt}
\caption{Visualization results for a {\it melanoma} sample.}
\label{fig:vis_mel}
\vspace{-10pt}
\end{figure}

\begin{figure}
\begin{minipage}[b]{1.0\linewidth}
  \centering
  \centerline{\includegraphics[width=5cm]{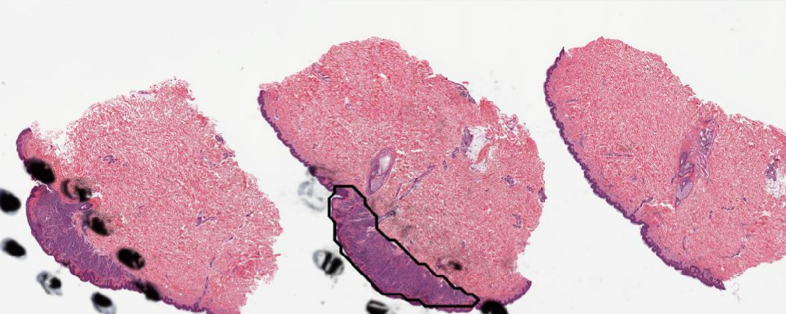}}
  \centerline{(a) Boundary of the largest predicted ROI region}\medskip
\end{minipage}
\begin{minipage}[b]{.48\linewidth}
  \centering
  \centerline{\includegraphics[width=5cm]{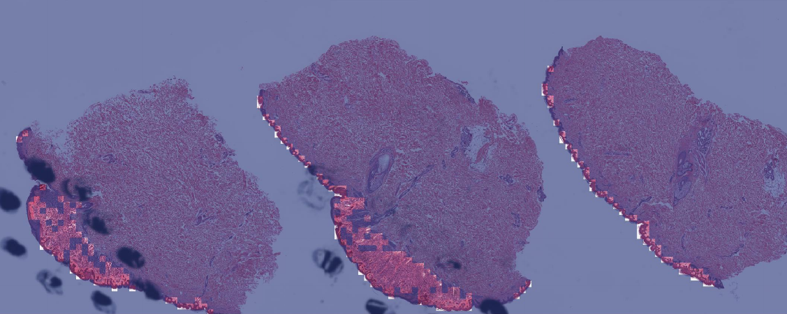}}
  \centerline{(b) Overlay}\medskip
\end{minipage}
\hfill
\begin{minipage}[b]{0.48\linewidth}
  \centering
  \centerline{\includegraphics[width=5cm]{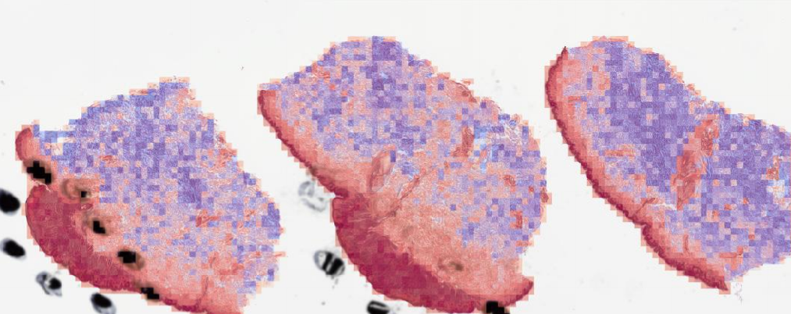}}
  \centerline{(c) Heatmap}\medskip
\end{minipage}
\vspace{-10pt}
\caption{Visualization results for a {\it nevus} sample.}
\label{fig:vis_nev}
\vspace{-10pt}
\end{figure}

\textit{Explanations for the figures}. Figure~\ref{fig:10} shows one representative sample of annotations (black dot regions) as well as without annotations. Visualization maps of two samples from the melanocytic skin tumor dataset are shown in figure~\ref{fig:vis_mel} and figure~\ref{fig:vis_nev}. Figure~\ref{fig:vis_mel} shows the three types of figures of the largest ROI region from melanoma found by the proposed method. Figure~\ref{fig:vis_nev} is the same for the nevus sample. The overlap map highlights top-ranked patches in a WSI and masks other area with a transparent blue color. The percentage of highlighted patches equals $\beta$ (the annotated ratio). The boundary map shows the boundary of the largest ROI cluster based on the highlighted patches, where the highlighted patches are clustered by OPTICS algorithm from \citet{ankerst1999optics}. The last one is a heatmap where red covers regions that have high predicted scores and blue covers regions that have low predicted scores.

\end{document}